\documentclass[sigconf,nonacm]{acmart}

\usepackage[T1]{fontenc}
\usepackage[utf8]{inputenc}
\usepackage{amsmath}
\usepackage{booktabs}
\usepackage{tabularx}
\usepackage{array}
\usepackage{graphicx}
\usepackage{url}
\usepackage{enumitem}

\renewcommand\footnotetextcopyrightpermission[1]{}
\newcolumntype{Y}{>{\centering\arraybackslash}X}
\newcolumntype{L}{>{\raggedright\arraybackslash}X}

\title{Efficient Partitioning Method of Large-Scale Public Safety Spatio-Temporal Data Based on Information Loss Constraints}

\author{Jie Gao}
\affiliation{%
  \institution{Beijing Key Laboratory of Intelligent Communication Software and Multimedia, School of Computer Science (National Pilot Software Engineering School), Beijing University of Posts and Telecommunications}
  \city{Beijing}
  \country{China}}

\author{Yawen Li}
\authornote{Corresponding author.}
\affiliation{%
  \institution{School of Economics and Management, Beijing University of Posts and Telecommunications}
  \city{Beijing}
  \country{China}}
\email{warmly0716@126.com}

\author{Zhe Xue}
\affiliation{%
  \institution{Beijing Key Laboratory of Intelligent Communication Software and Multimedia, School of Computer Science (National Pilot Software Engineering School), Beijing University of Posts and Telecommunications}
  \city{Beijing}
  \country{China}}

\author{Zeli Guan}
\affiliation{%
  \institution{Beijing Key Laboratory of Intelligent Communication Software and Multimedia, School of Computer Science (National Pilot Software Engineering School), Beijing University of Posts and Telecommunications}
  \city{Beijing}
  \country{China}}

\begin{abstract}
The storage, management, and application of massive spatio-temporal data are widely used in practical scenarios, including public safety. However, due to the unique spatio-temporal distribution characteristics of real-world data, existing methods still face limitations in preserving spatio-temporal proximity and achieving load balancing in distributed storage. This paper proposes an efficient partitioning method for large-scale public safety spatio-temporal data based on information loss constraints, named IFL-LSTP. The model combines a spatio-temporal partitioning module (STPM) and a graph partitioning module (GPM). STPM reduces the scale of data under a predefined information-loss threshold, while GPM uses graph representation learning to obtain balanced graph partitions. Experiments on multiple real-world datasets show that IFL-LSTP can reduce data scale, shorten graph model training time, preserve spatio-temporal proximity, and improve load-balancing effectiveness.
\end{abstract}


\begin{document}
\maketitle

\section{Introduction}
In recent years, the scale of spatio-temporal data has grown rapidly. Distributed data management systems and spatio-temporal data analytics platforms are therefore widely used to store, organize, and analyze massive data streams \cite{xiao2022lecf,alam2022stsurvey}. Public safety data usually contains uneven spatial aggregation and strong temporal correlations. These properties make it challenging to partition large-scale, unevenly distributed data while preserving spatio-temporal proximity and maintaining load balancing. Large-scale path query, POI retrieval, and social media mining studies show that heterogeneous spatio-temporal and social signals require efficient indexing, representation, and retrieval mechanisms \cite{li2022distributedpath,huang2021hgamn,kou2018hashtag}. Scientific and technological information retrieval methods also show that heterogeneous and cross-media data require careful representation before efficient retrieval and analysis can be performed \cite{li2022smcr}. Cross-modal federated retrieval is less directly related to spatio-temporal partitioning, but it provides a useful privacy-preserving reference for distributed public safety data services \cite{li2024fedcmr}.

To address the above problems, this paper proposes an efficient partitioning method for large-scale public safety spatio-temporal data based on information loss constraints, called IFL-LSTP. IFL-LSTP consists of two main modules: the spatio-temporal partitioning module (STPM) and the graph partitioning module (GPM). STPM reduces the grid scale under a prescribed information-loss threshold, and GPM performs balanced graph partitioning by considering both normalized edge cut and partition load.

The contributions of this work are summarized as follows.
\begin{itemize}[leftmargin=*]
  \item We propose IFL-LSTP, an efficient partitioning method for large-scale public safety spatio-temporal data, which combines STPM and GPM to maintain spatio-temporal proximity and load balancing.
  \item STPM iteratively reduces data size under a predefined information-loss threshold without requiring the final number of partitions to be specified in advance.
  \item GPM adopts graph embedding and graph neural network techniques to design a loss function that jointly considers minimum normalized edge cut and partition load balancing.
\end{itemize}

\section{Related Work}
\subsection{Spatio-Temporal Data Management and Distributed Learning}
Large-scale spatio-temporal data management is closely related to distributed storage, multimodal path query, and heterogeneous data retrieval. Existing studies on distributed multimodal path queries, POI retrieval, vehicle fuel-consumption prediction, and large graph traversal provide useful background for managing high-volume data with spatial, temporal, and relational characteristics \cite{li2022distributedpath,huang2021hgamn,li2022fuel,shao2021memorywalk}. Distributed consensus and tracking studies also provide general references for processing observations across networked systems \cite{lin2009averageconsensus,meng2013tracking,li2014phd}. Federated learning with stochastic quantization and FedSIN are relevant to privacy-aware or communication-efficient distributed representation learning, especially when data cannot be centralized \cite{li2022stochasticquantization,li2026fedsin}.

\subsection{Graph Partitioning and Graph Representation}
Graph partitioning and graph representation learning provide the theoretical and technical basis for the graph partitioning module. Approximate graph partitioning, graph embedding, graph coarsening, and matrix operations such as the Hadamard product are closely related to balanced graph-cut optimization \cite{nazi2019gap,xu2021graphembedding,luo2022netclusting,horn2020hadamard}. Graph convolutional networks and GraphSAGE provide common graph representation mechanisms, while federated graph neural networks extend graph learning to cross-graph and distributed settings \cite{kipf2016gcn,hamilton2017graphsage,guan2021fedgnn}. Studies on community detection, heterogeneous graph attention, self-supervised heterogeneous graph learning, and T2-GNN further indicate that complex real-world graphs often require robust representation learning under heterogeneous or incomplete structures \cite{yang2016modularity,hu2019hgat,jin2022reciprocalcontrastive,huo2023t2gnn}.

\subsection{Heterogeneous Data Representation and Knowledge Services}
Public-safety data services may need to integrate heterogeneous observations, semantic retrieval, decision support, and knowledge services. Heterogeneous latent topic discovery, incomplete multi-view multi-label learning, and semantic-similarity hypergraph representation learning provide useful references for representing heterogeneous or higher-order relations \cite{li2021hltd,ou2024vcit,li2026ssahgc}. For text-oriented public-safety and social-media content, fake-news detection with deep Markov random fields and sarcasm detection with knowledge-enhanced entity and relationship understanding show that relation-aware semantic modeling can support reliability analysis of noisy online signals \cite{dong2023dmrf,wang2025sarcasm}. Multi-view scholar clustering is not a direct spatio-temporal partitioning baseline, but it is related to dynamic entity grouping and portrait construction \cite{li2023mvsc}. Recommendation and retrieval-oriented studies, including filter-enhanced MLP, self-supervised graph co-training, Tucker-decomposition dataset distillation, and RetroMAE, are less central to this partitioning task but provide references for efficient sequence modeling, graph interaction modeling, and retrieval-oriented representation learning \cite{zhou2022fmlp,xia2021graphcotraining,zhang2025td3,xiao2022retromae}. Interpretable machine learning and human-centered innovation studies are also useful for future public-safety decision services that require explainability and human-oriented evaluation \cite{li2019interpretable,zhou2020creativeentrepreneur}. Other intelligent-system studies, including crowd counting, business computing, Kalman filtering, and neural learning systems, provide auxiliary background for data-driven intelligent applications \cite{wei2019crowd,li2018business,li2017tobit}.

\section{Methodology}
\subsection{Problem Definition}
\textbf{Definition 1. Information loss.} The information difference between the initial input dataset and the result after STPM partitioning is called information loss (IFL). For a grid with $n$ cells and $s$ spatio-temporal attributes, IFL is calculated as
\begin{equation}
\mathrm{IFL}=\frac{1}{n}\sum_{i=1}^{n}\sum_{k=1}^{s}\frac{|d_i(k)-d'_i(k)|}{d_i(k)},
\end{equation}
where $d_i(k)$ denotes the original value of attribute $k$ in cell $i$, and $d'_i(k)$ denotes the representative value of the corresponding attribute after STPM partitioning.

\textbf{Definition 2. Minimum normalized edge cut.} Given an undirected graph $G=(V,E)$, the graph can be divided into $g$ disjoint node sets $S_1,S_2,\ldots,S_g$. The minimum normalized edge cut is defined as
\begin{equation}
\mathrm{MNcut}(S_1,S_2,\ldots,S_g)=\sum_{k=1}^{g}\frac{\mathrm{cut}(S_k,\bar S_k)}{\mathrm{vol}(S_k,V)},
\end{equation}
where $\mathrm{vol}(S_k,V)$ denotes the total degree of the nodes belonging to set $S_k$ in graph $G$. This definition follows the normalized graph-cut objective used in the graph partitioning module.

\subsection{Spatio-Temporal Partitioning Module}
\begin{figure*}[t]
\centering
\includegraphics[width=0.95\textwidth]{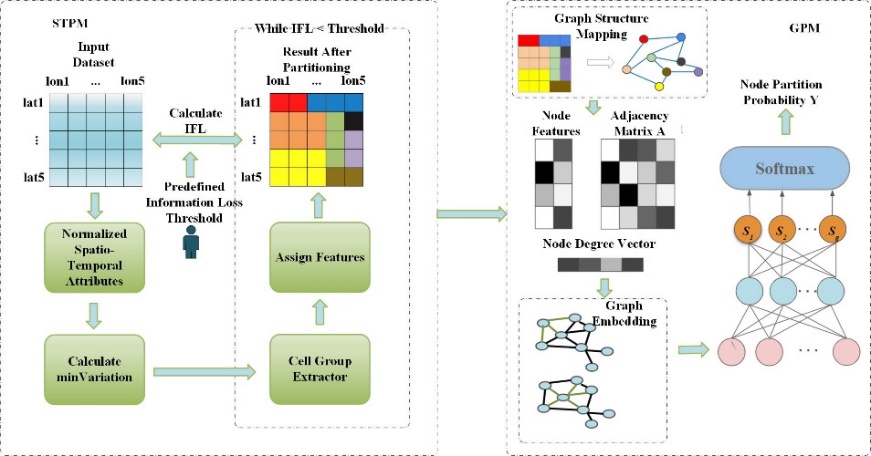}
\caption{Illustration of IFL-LSTP. The left part shows STPM, which controls information loss during spatio-temporal grid reduction. The right part shows GPM, which performs graph partitioning through graph representations and a load-aware objective.}
\label{fig:ifl_lstp}
\end{figure*}

As shown in Fig.~\ref{fig:ifl_lstp}, STPM first normalizes the spatio-temporal attributes of the dataset and pre-calculates the minimum neighboring attribute difference values. The cell-group extractor traverses the grid to find rectangular groups of neighboring cells. The attribute differences within each group should be less than or equal to the minimum neighboring attribute difference threshold. Once all eligible cell groups are found, each group is treated as a single cell in the next iteration.

The feature allocator generates representative values for each extracted cell group. STPM defines an allocation selection parameter $\lambda$, whose values include average and mode. The former assigns the average value of the spatio-temporal attributes in the group, while the latter assigns the most frequent value. After features are assigned, STPM calculates the information loss between the current partitioning result and the original input grid, and determines whether the result is below the predefined threshold $\theta$. If the threshold is satisfied, the reduced grid is passed to GPM; otherwise, STPM continues the iteration.

\subsection{Graph Partitioning Module}
The output of GPM is a matrix $Y\in \mathbb{R}^{n\times g}$, where $Y_{ik}$ denotes the probability that node $v_i$ is assigned to partition $S_k$. Based on the adjacency matrix $A$ of the graph, the minimum normalized edge cut can be relaxed as
\begin{equation}
\mathbb{E}[\mathrm{MNcut}(S_1,\ldots,S_g)] = \sum_{k=1}^{g}\frac{Y_k^{\top}(\mathbf{1}\mathbf{1}^{\top}-A)Y_k}{Y_k^{\top}D},
\end{equation}
where $D$ is the node degree vector. For a graph with $g$ partitions and load $f_i$ on node $v_i$, the expected average load is $\sum_{i=1}^{n}f_i/g$. The loss function of GPM is therefore specified as
\begin{equation}
\mathcal{L}_{\mathrm{GPM}}=\mathbb{E}[\mathrm{MNcut}(S_1,\ldots,S_g)] + \sum_{k=1}^{g}\left(e_k^{\top}F-\frac{1}{g}\sum_{i=1}^{n}f_i\right)^2 .
\end{equation}
This loss jointly optimizes graph-cut quality and load balance.

GPM uses graph embedding to learn graph structures. Specifically, the input contains the adjacency matrix $A$, the node degree vector $D$, and node features $X$. The model first applies a two-layer graph convolutional structure to extract node-level representations and then uses an inductive neighborhood aggregation module to generate high-dimensional node embeddings. Finally, the learned node embeddings are fed into a fully connected layer and SoftMax to output the probability that each node belongs to each partition.

\section{Experiments}
\subsection{Experimental Data Preparation}
To verify the effectiveness of IFL-LSTP, experiments are conducted on three real-world datasets: the house-sales dataset, the GLONASS+112 dataset, and a public-safety dataset. The GLONASS+112 dataset is an emergency dataset, from which 100,000 emergency events are randomly sampled to form the mGLONASS dataset for evaluation \cite{dagaeva2019bigst}. The public-safety dataset is crawled and organized from a microblogging platform. In addition, two grid sizes, 100k $(315\times318)$ and 36k $(191\times193)$, are used to evaluate data size reduction under information loss thresholds of 0.05, 0.1, and 0.15.

\subsection{Evaluation of Data Size Reduction}
\begin{figure*}[t]
\centering
\begin{tabular}{ccc}
\includegraphics[width=0.30\textwidth]{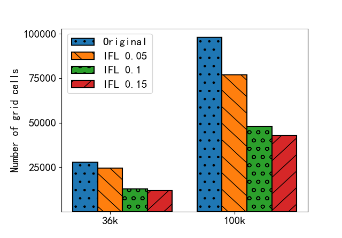} &
\includegraphics[width=0.30\textwidth]{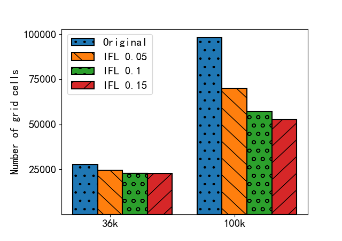} &
\includegraphics[width=0.30\textwidth]{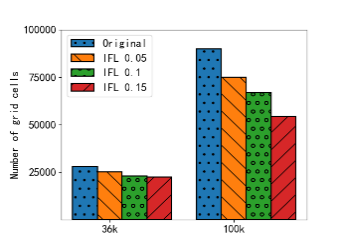} \\
(a) house-sales & (b) mGLONASS & (c) public-safety \\
\includegraphics[width=0.30\textwidth]{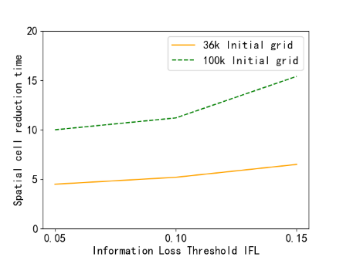} &
\includegraphics[width=0.30\textwidth]{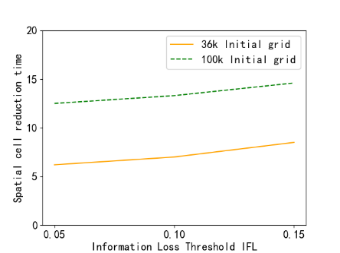} &
\includegraphics[width=0.30\textwidth]{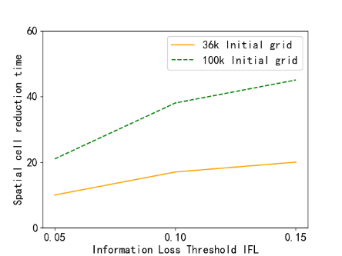} \\
(d) house-sales & (e) mGLONASS & (f) public-safety \\
\end{tabular}
\caption{Evaluation of cell reduction performance of IFL-LSTP with various IFL values on all datasets.}
\label{fig:cell_reduction}
\end{figure*}

Figure~\ref{fig:cell_reduction} shows the effect of data size reduction on the three datasets with different information-loss thresholds. Subfigures (a)--(c) show that IFL-LSTP can reduce the number of grid cells by about 25\% when the information loss is only 0.05. When the IFL threshold increases to 0.1 and 0.15, the number of grid cells can be further reduced up to about 36\%. However, the reduction rate gradually slows as the threshold increases.

Subfigures (d)--(f) show grid cell reduction times under different IFL thresholds. The reduction time increases with the IFL threshold and initial grid granularity because finer grids require STPM to process more cell groups. When the threshold becomes larger, STPM performs more iterations and therefore requires more running time.

\subsection{Evaluation of Spatio-Temporal Proximity}
To verify the spatio-temporal proximity of IFL-LSTP partitioning results, we compare IFL-LSTP with the spatio-temporal hierarchical indexing method, TrajSpark, and LE \cite{zhao2019stindex,zhang2017trajspark,xia2018le}. The results are shown in Table~\ref{tab:proximity}. IFL-LSTP achieves the best performance at IFL=0.15 on the house-sales and public-safety datasets, reaching 85.79\% and 81.38\% spatio-temporal proximity, respectively. On mGLONASS, IFL-LSTP with IFL=0.15 reaches 82.14\%, which is close to the strongest baseline result. Compared with TrajSpark, IFL-LSTP with IFL=0.15 improves spatio-temporal proximity by 5.76\% on average over all datasets. This demonstrates that IFL-LSTP can maintain proximity while reducing data scale.

\begin{table*}[t]
\caption{Comparison results of spatio-temporal proximity.}
\label{tab:proximity}
\begin{tabularx}{\textwidth}{@{}p{0.20\textwidth}L Y@{}}
\toprule
Datasets & Methods & Spatio-temporal proximity $p$/\% \\
\midrule
house-sales & TrajSpark & 80.43 \\
 & Spatio-temporal hierarchical indexing & 84.67 \\
 & LE & 85.06 \\
 & IFL-LSTP (IFL=0.05) & 83.28 \\
 & IFL-LSTP (IFL=0.1) & 85.54 \\
 & IFL-LSTP (IFL=0.15) & \textbf{85.79} \\
\midrule
mGLONASS & TrajSpark & 78.76 \\
 & Spatio-temporal hierarchical indexing & 81.33 \\
 & LE & \textbf{82.39} \\
 & IFL-LSTP (IFL=0.05) & 79.40 \\
 & IFL-LSTP (IFL=0.1) & 80.97 \\
 & IFL-LSTP (IFL=0.15) & 82.14 \\
\midrule
public-safety & TrajSpark & 76.53 \\
 & Spatio-temporal hierarchical indexing & 78.89 \\
 & LE & 80.12 \\
 & IFL-LSTP (IFL=0.05) & 77.74 \\
 & IFL-LSTP (IFL=0.1) & 80.06 \\
 & IFL-LSTP (IFL=0.15) & \textbf{81.38} \\
\bottomrule
\end{tabularx}
\end{table*}

\subsection{Training Time and Load Balancing}
\begin{figure*}[t]
\centering
\includegraphics[width=0.52\textwidth]{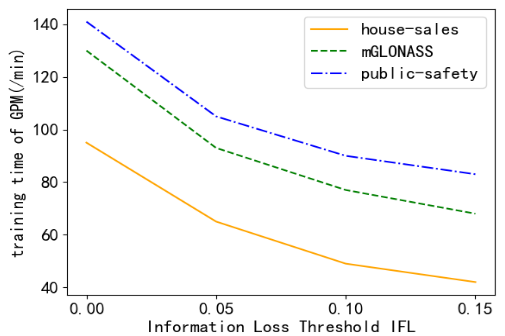}
\caption{Analyzing training time of GPM with various IFL values on all datasets.}
\label{fig:training_time}
\end{figure*}

Figure~\ref{fig:training_time} analyzes the training time of GPM under different IFL thresholds. IFL-LSTP with IFL=0.05 can reduce GPM training time by about 25\%--32\%. Increasing IFL to 0.1 and 0.15 reduces grid cells more aggressively, but does not always reduce training time proportionally. Therefore, a trade-off must be made between reducing training time and preserving model accuracy.

Table~\ref{tab:loadbalance} reports the evaluation of load balancing using edge cut rate $\lambda$ and overall imbalance $\mathrm{ubd}$. IFL-LSTP improves the graph partitioning indices compared with the classical partitioning methods Metis and LDG \cite{luo2022netclusting,pacaci2019streaming}. This is because IFL-LSTP first performs spatio-temporal partitioning and then considers the influence of spatio-temporal distribution on load balancing. The edge cut rate and imbalance gradually decrease as IFL increases, but the difference between IFL=0.1 and IFL=0.15 is relatively small. This further verifies the need for a balanced threshold selection strategy.

\begin{table*}[t]
\caption{Evaluation of load balancing effectiveness.}
\label{tab:loadbalance}
\begin{tabularx}{\textwidth}{@{}p{0.20\textwidth}L YY@{}}
\toprule
Datasets & Methods & $\lambda$ & $\mathrm{ubd}$/\% \\
\midrule
house-sales & Metis & 0.19 & 1.76 \\
 & LDG & 0.13 & 1.69 \\
 & IFL-LSTP (IFL=0.05) & 0.15 & 1.68 \\
 & IFL-LSTP (IFL=0.1) & 0.11 & 1.23 \\
 & IFL-LSTP (IFL=0.15) & \textbf{0.09} & \textbf{1.12} \\
\midrule
mGLONASS & Metis & 0.26 & 2.35 \\
 & LDG & 0.22 & 2.17 \\
 & IFL-LSTP (IFL=0.05) & 0.23 & 2.21 \\
 & IFL-LSTP (IFL=0.1) & 0.17 & 1.79 \\
 & IFL-LSTP (IFL=0.15) & \textbf{0.16} & \textbf{1.72} \\
\midrule
public-safety & Metis & 0.28 & 2.63 \\
 & LDG & 0.25 & 2.54 \\
 & IFL-LSTP (IFL=0.05) & 0.25 & 2.52 \\
 & IFL-LSTP (IFL=0.1) & 0.20 & 2.07 \\
 & IFL-LSTP (IFL=0.15) & \textbf{0.18} & \textbf{1.99} \\
\bottomrule
\end{tabularx}
\end{table*}

\section{Conclusion}
This paper proposes IFL-LSTP, an efficient partitioning method for large-scale public safety spatio-temporal data based on information loss constraints. IFL-LSTP combines STPM and GPM to reduce data scale, preserve spatio-temporal proximity, and achieve load-balanced graph partitioning. Experiments on three real-world datasets verify the effectiveness of the method in data reduction, training-time reduction, proximity preservation, and load balancing. Future work may extend IFL-LSTP to privacy-preserving distributed learning, incomplete graph data, and interpretable public-safety decision services.

\begin{acks}
This work was supported by the National Natural Science Foundation of China (62192784, U22B2038, 62172056, 62272058).
\end{acks}

\bibliographystyle{jiegao_custom_unsrt}
\bibliography{references}

\end{document}